# Adversarial Attacks on Reinforcement Learning based Energy Management Systems of Extended Range Electric Delivery Vehicles

Pengyue Wang, Yan Li, Shashi Shekhar, *Fellow, IEEE,* William F. Northrop*

*Abstract*— Adversarial examples are firstly investigated in the area of computer vision: by adding some carefully designed "noise" to the original input image, the perturbed image that cannot be distinguished from the original one by human can fool a well-trained classifier easily. In recent years, researchers also demonstrated that adversarial examples can mislead deep reinforcement learning (DRL) agents on playing video games using image inputs with similar methods. However, although DRL has been more and more popular in the area of intelligent transportation systems, there is little research investigating the impacts of adversarial attacks on them, especially algorithms that do not use images as inputs. In this work, we investigated several fast methods to generate adversarial examples to significantly degrade the performance of a well-trained DRL-based energy management system of an extended range electric delivery vehicle. The perturbed inputs are low dimensional state representations and close to the original input quantified by $L_1$, $L_2$ or $L_\infty$ norms. Our work shows that to apply DRL agents on real-world transportation systems, adversarial examples in the form of cyber-attack should be considered carefully, especially for applications that may lead to serious safety issues.

## I. Introduction

Reinforcement learning (RL) algorithms are used to solve sequential decision-making problems without a model of the environment or the studied system [1]. If a model is available, dynamic programming (DP) can be used to solve the problem with an optimal solution [2]. Traditional value-based RL algorithms use tables or linear models to represent action-value functions to solve small-scale problems. The features used for the linear models are usually designed by experts with strong domain knowledge. In recent years, with the rapid development of deep learning in classification and regression [3], deep neural networks (NNs) are also introduced into RL as function approximators. The NNs can extract useful features from raw input images or low dimensional state representations without a strong domain knowledge. With powerful function approximators, significant upgrades of computation hardware as well as various techniques to stable the training process, huge successes have been made in problems with a much larger scale, like video game playing [4][5] and continuous control of complex physical systems [6].

Many applications in the area of transportation involve a process of sequential decision-making, and usually the studied transportation systems are difficult to model accurately. Consequently, deep RL (DRL) algorithms have been adapted to many of them such as traffic light cycle control [7][8], autonomous driving [9][10] or energy management systems (EMS) of hybrid vehicles [11][12]. With emerging technologies like Vehicle-to-Vehicle and Vehicle-to-Cloud communications, DRL will be more widely investigated in the area of Intelligent Transportation Systems (ITS).

Robustness is one of the most important factors that needs to be considered when implementing a new system into the transportation area. However, it is found for a well-trained NN-based classifier, it is possible to add a small crafted perturbation to the original image so that the classifier would misclassify the perturbed image, even the difference are not discernable by human [13]. In [14], a method called Fast Gradient Sign Method (FGSM) is introduced to create such adversarial examples. It requires small computation resources and achieves very good results. In addition, they found one adversarial example created for a specific model can also be misclassified by other models with different architectures for the same task. Further, it is shown that the adversarial examples are still effective even they are perceived through camera in a real-world problem setting [15].

As NNs are also the core components of DRL algorithms, researchers have investigated the impact of adversarial examples under several simulation settings. In [16], FGSM and its two variations with different norm constraints were applied to degrade the performance of a DRL agent on playing video games using image inputs. Also, adversarial examples were compared with random noise under the same simulation environment in [17]. It was shown that adversarial examples were much more effective in misleading the agent. In [18], a more complex and computational expensive approach involving a generative model and a planning algorithm was introduced to lure the agent to designated target states. In [19], a systematic characterization of adversarial attacks was shown. They also performed comprehensive experiments on video game playing as well as physical system control which uses low dimensional state representation. The same physical system control simulation environment was also used in [20]. They developed adversarial robust policy learning algorithm to help the agent perform better at test time under the gap between simulation domain and physical domain caused by random noise or adversarial noise.

In this work, we investigate the impacts of adversarial examples generated by FGSM related methods on a well-behaved DRL-based EMS. The EMS is for an extended range electric vehicle (EREV) used for package delivery. The simulation is based on recorded real-world historical delivery trips and an accurate vehicle model. This study is among the

*Corresponding author
The information, data, or work presented herein was funded in part by the Advanced Research Projects Agency-Energy (ARPA-E) U.S. Department of Energy, under Award Number DE-AR0000795.

P. Wang, Y. Li, S. Shekhar and W. F. Northrop are with University of Minnesota, Minneapolis, MN 55455. (email: wang6609@umn.edu; lixx4266@umn.edu; shekhar@umn.edu; wnorthro@umn.edu)

first to investigate the adversarial examples on DRL algorithms in the area of ITS, where robustness and safety are of great importance. Different attacks under different assumptions are performed, and Fig.1 illustrates the attack process. It is shown that adversarial attacks can degrade the performance of the DRL-based EMS significantly, either causing the EREV using too much fuel, or leading it running out of battery during the delivery trip.

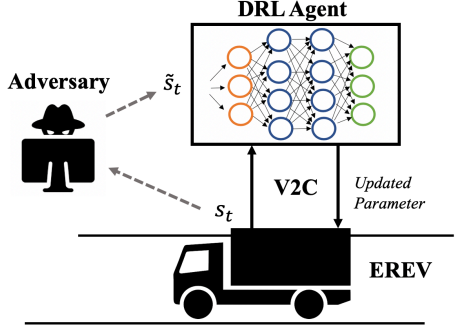

Fig. 1. Illustration of the adversarial attacks. The low dimensional state representation $s_t$ provided by the EREV is processed by the adversary before sending to the DRL agent on the cloud.

## II. REINFORCEMENT LEARNING BASED ENERGY MANAGEMENT SYSTEMS FOR EREV

### A. Energy Management System Introduction

The powertrain configuration of the studied EREV is shown in Fig. 2. The high-capacity battery is the main energy source of the vehicle. The internal combustion engine serves as a range-extender which is used to charge the battery through the generator according to the EMS. There is no mechanical connection between the engine output shaft and the drive shaft; the motive power of the vehicle is provided solely by the electric motor.

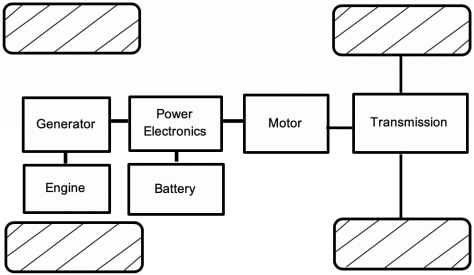

Fig. 2. Powertrain configuration of the EREV

Ideally, to achieve high fuel efficiencies and reduce on-road emissions, no fuel should be used to charge the battery if the delivery trip is short, which does not exceed the all-electric range (AER) of the EREV. For longer trips that require energy from the range-extender, the goal is to use as less fuel as possible and achieve a target end state of charge (SOC) at the end of the trip, which should also be the lowest SOC during the whole trip. During the delivery trip, if the real-time measured SOC is lower than a calculated reference value, the engine will be turned on at a predefined high efficiency speed and load condition. The reference value is calculated as:

$$SOC_{ref} = 100\% \times \left(1 - 0.9 \frac{d_t}{L_{set}}\right), \quad (1)$$

where $d_t$ represents the distance the vehicle has traveled, and $L_{set}$ is the energy-compensated expected trip distance. The value of 0.9 comes from the setting that the target end SOC is 10% in this work as well as in the studied real-world delivery fleet. In addition, if $SOC_{ref}$ is higher than 60%, it is set to be 60% to prevent fuel use in very short trips.

The physical meaning behind $SOC_{ref}$ is how much energy is expected to be left when the vehicle has traveled for $d_t$ given the value of $L_{set}$. The value of $L_{set}$ can be understood intuitively: the higher the value of $L_{set}$, the more fuel would be used to charge the battery as the EMS considers there is a long way to go. A good value of $L_{set}$ can help the vehicle achieve high fuel efficiency and always have enough electric energy ($SOC > 10\%$). However, even for the same vehicle running on a certain delivery area, trip statistics like distance, energy intensity and GPS trajectory vary day-to-day due to factors like delivery demand, weather and traffic condition. The distribution of distance and energy intensity of the studied EREV is shown in Fig. 3.

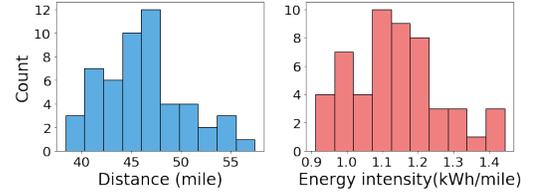

Fig. 3. Distance and energy intensity distribution of the studied EREV

To optimize the value of $L_{set}$ adaptively during ongoing delivery trips, a RL-based EMS has been developed. It uses real-time information provided by the vehicle and an agent trained on historical delivery trips of the vehicle. The updated $L_{set}$ value can be provided to the running vehicle through a Vehicle-to-Cloud (V2C) connectivity. The example trips using the RL-based EMS are shown in Fig. 4 and 5 [11].

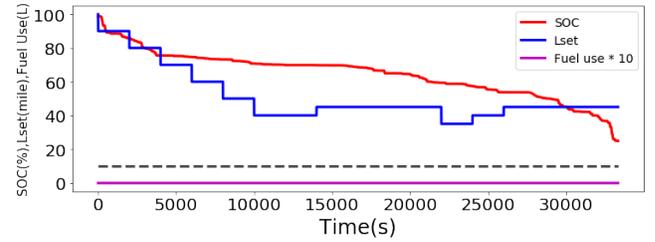

Fig. 4. Performance of the studied DQN on a 40-mile delivery trip

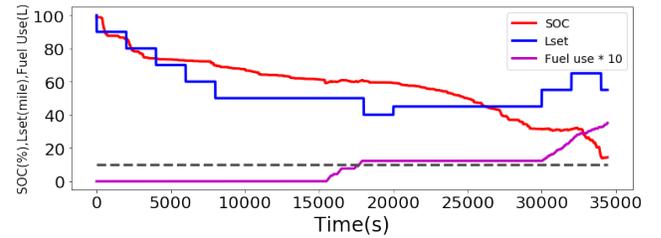

Fig. 5. Performance of the studied DQN on a 50-mile delivery trip

### B. Reinforcement Learning Algorithms

This work involves two model-free value-based DRL algorithms. In this part, the formulation of the RL problem is first discussed and then the two DRL algorithms are briefly introduced.

*1) Formulation*

The state space consists of the real-time information that is available during the delivery trip, including traveled time, distance, current SOC, fuel use, GPS position and the current $L_{set}$ setting. Each state in the state space can be represented by a vector consisting of seven entries:

$$s_t = [t_{travel}, d, SOC, f, x, y, L_{set}], \quad (2)$$

where all the dimensions are scaled to a range of [0,1].

The action space is a set of predefined $L_{set}$ changes:

$$a_t \in [-10, -5, 0, +5, +10]. \quad (3)$$

The environment $p(s_{t+1}|s_t, a_t)$ is approximated by a simplified vehicle model and 52 historical delivery trips of the studied vehicle with a distance range of 38 to 57 miles and an energy intensity range of 0.91 to 1.44 kWh/miles.

The reward function is defined as:

$$r_t = r_f t_{f,t} + r_{SOC} t_{SOC,t} + r_{a,t} + r_c. \quad (4)$$

The first term penalizes fuel use during the delivery trip. The magnitude is proportional to the engine running time with a coefficient of $-0.001$. The second term penalizes the condition of $SOC < 10\%$ and its magnitude is proportional to the time under that condition with a coefficient of $-0.060$. The third term equals to $-0.020$ if the chosen action is not 0 and equals to 0 otherwise. It can lead to a more efficient policy with less frequent $L_{set}$ changes. The last term is only given at the terminal state of the task. It compensates for the negative reward caused by the necessary fuel use that keeps the SOC always higher than 10% during the trip, and this can only be calculated when the trip is finished.

*2) Deep Q-Network*

Deep Q-Network (DQN) is one of the most popular DRL methods in recent years and there are many versions of it with different incremental improvements [4][5]. The key concept for all of them is the action-value function defined as:

$$Q_\pi(s, a) = E_\pi[G_t | s_t = s, a_t = a], \quad (5)$$

where $G_t$ is the long-term return with discount factor $\gamma$:

$$G_t = r_t + \gamma r_{t+1} + \gamma^2 r_{t+2} + \cdots + \gamma^{T-1-t} r_{T-1}. \quad (6)$$

$Q_\pi(s, a)$ represents the expected long-term return the agent can achieve at the defined environment at timestep $t$ if it takes action $a$ at state $s$ and then following policy $\pi$. Intuitively, it quantifies how good is action $a$ at state $s$.

The core of DQN-based methods is to use NNs as function approximators for the optimal action-value function:

$$Q^*(s, a) = max_\pi Q_\pi(s, a), \quad (7)$$

which is the highest action-value that can be achieved for all state-action pairs under all possible policies. The parametrized action-value function $Q(s, a; \psi)$ is also called Q-network.

The policy can be derived by acting greedily with respect to $Q(s, a; \psi)$:

$$\pi(s) = argmax_a Q(s, a; \psi). \quad (8)$$

*3) Implicit Quantile Network*

Implicit Quantile Network (IQN) [21] is a distributional DRL method. Although it is also a value-based method, instead of estimating the action-value directly, it models the full return distribution by a NN parametrized by $\phi$. In other words, for a give state-action pair, DQN will output the expected long-term return as introduced above; the IQN will output $K$ samples of the long-term return according to the implicitly modeled return distribution. The action-value can be estimated from the samples:

$$Q(s, a; \phi) = E_{\tau \sim U([0,1])}[Z_\tau(s, a; \phi)] \approx \frac{1}{K} \sum_{i=1}^{K} Z_{\tau_i}(s, a; \phi), \quad (9)$$

where index $i$ represents the $ith$ sample from the IQN and $\tau$ is another input to the IQN, sampled from a uniform distribution $U([0,1])$. $Z_{\tau_i}(s, a; \phi)$ is one sample of the possible future return whose expected value equals to the action-value.

## III. ADVERSARIAL ATTACKS METHODS

In this section, several fast methods to generate adversarial examples used in this work are introduced. They are categorized as white box methods and black box methods according to the information known to the attacker.

### A. White box

In this condition, it is assumed that the attacker has full access to the target agent including network structure, parameter weights as well as all the related training details. FGSM [14] is used with two variations [16].

The main idea of the original FGSM is straightforward. Given a loss function $J(\theta, x, y)$ where $\theta$ represents the weights in the NN-based classifier, $x$ represents the input image with $d$ dimensions and $y$ is the true label corresponding to $x$, a small perturbation is added to the original input $x$ so that the loss function $J(\theta, \tilde{x}, y)$ on the perturbed image $\tilde{x}$ is increased as much as possible. This is achieved by:

$$\tilde{x} = x + \varepsilon sign[\nabla_x J(\theta, x, y)], \quad (10)$$

which is derived by linearizing the cost function around the current parameters $\theta$, and solving a $L_\infty$ constraint optimization problem. For a more detailed and comprehensive analysis, we recommend the reader to the original papers [14][15]. In addition, the $L_1$ and $L_2$ norm constrained form is also included [16]. For $L_1$, we maximally perturb the dimension $i$ that has the highest gradient magnitude:

$$\tilde{x} = x + \varepsilon d \cdot e_i, \quad (11)$$

where $e_i$ is a $d$ dimensional vector whose $ith$ component is 1 and 0 otherwise. For $L_2$ constraint, the perturbed input is:

$$\tilde{x} = x + \varepsilon \sqrt{d} \frac{\nabla_x J(\theta, x, y)}{\|\nabla_x J(\theta, x, y)\|_2}. \quad (12)$$

In the RL setting, to get the perturbed $\tilde{s}_t$ from $s_t$, the action-value vector is first calculated from the DRL agent. Then, it is transformed into probabilities $y_p$ by the softmax function. The label $y$ is obtained by replacing the position of the highest entry in $y_p$ with 1 and 0 otherwise. The loss $J(\theta, s_t, y)$ is then calculated as the cross-entropy between $y_p$ and $y$. Cross-entropy loss is used for classification and in the RL setting, it is assumed the target agent performs well and what we want is to make it choose another action instead of the action based on the original input ("misclassify actions").

## B. Black box

### 1) Finite Difference Method

As the NN structure and weights are unknown in the black box condition, the gradient $\nabla_{s_t} J(\theta, s_t, y)$ cannot be calculated directly. One method is to estimate it by finite difference (FD) method [19]. To apply it on an input with $d$ dimensions, it requires querying the target agent $2d$ times. The $ith$ component of the estimated gradient $\nabla_{s_t} \tilde{J}(\theta, s_t, y)$ is:

$$\frac{J(\theta, s_t + \delta e_i, y) - J(\theta, s_t - \delta e_i, y)}{2\delta}, \quad (13)$$

$\delta$ is a hyperparameter that controls the accuracy. After $\nabla_{s_t} \tilde{J}(\theta, s_t, y)$ is calculated, it can be used as $\nabla_{s_t} J(\theta, s_t, y)$. This method can be computationally expensive if the input dimension is high, for example, an image. In this work, the input is a low dimensional state representation so that the computation burden is insignificant.

### 2) Transferability Property

If the attacker is not allowed to query the target agent, but has access to the training environment, the transferability property of adversarial examples can be utilized. It is found that adversarial examples generated for one NN often works for another NN even the parameter weights and structures are different [14][16]. Therefore, the attacker can train an agent for the same task using the training environment and generate $\tilde{s}_t$ using its own agent for the unknown target agent.

We investigate two cases: transfer across policies and transfer across algorithms. For the first case, it is assumed that the attacker knows the target agent's used RL algorithm, NN structures and training hyperparameters. For the second case, the attacker only has access to the training environment. There are plenty of variations of experiments that can be done, for example, for the transfer across policies, the attacker may only know the RL algorithm used and nothing else. Although we only experiment with the described two special cases, it is enough to show initial results about the transferability property for this problem.

## IV. SIMULATION RESULTS

We applied the introduced methods on a DQN agent [11] on 52 recorded historical delivery trips. Two kinds of random noise are used as baselines. The first is random sign noise. A value with magnitude $\varepsilon$ is added to each dimension and the sign is randomly chosen from positive and negative with equal probabilities. The second kind of noise is uniform random noise. For each dimension, a value sampled from a uniform distribution $U([-\varepsilon, \varepsilon])$ is added to the original input.

The performance of the DQN under these two kinds of noise is shown in Fig. 6. For each value of $\varepsilon$, we run 10 experiments and get 10 average scores.

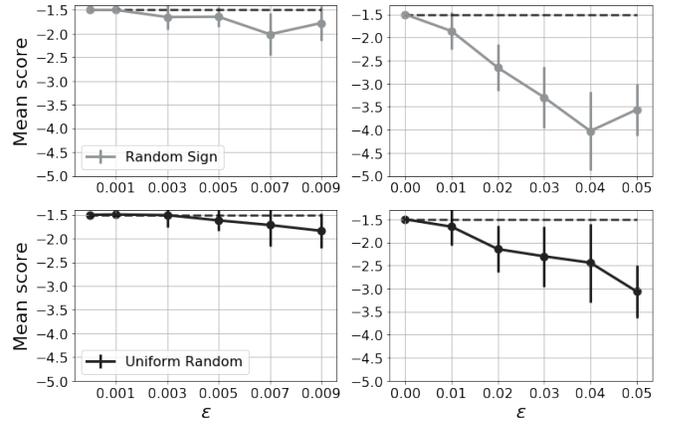

Fig. 6. Performance of the DQN under two kinds of random noise with different values of $\varepsilon$. The error bars indicate one standard deviation.

The performance of the DQN under FGSM and its two variations are summarized in Fig. 7. First, it can be observed that all the three methods are more effective than random noise. For the same magnitude of $\varepsilon$, the gradient-based adversarial examples can degrade the agent to much worse scores. On the other hand, to misguide the agent to a certain low level of performance, adversarial examples are much more similar to the original inputs. Second, it is obvious the $L_1$ constrained method performs best for most cases. However, one major downside is, it uses all budget $\varepsilon d$ to perturb one dimension so that it is much easier to be detected.

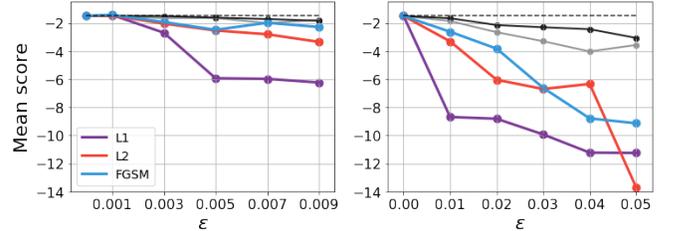

Fig. 7. Performance of the DQN under FGSM and its two variations

Fig. 8. compares the performance of FGSM under white box condition and its estimation with FD method under black box condition. $\delta$ is set to be 0.0001. It can be observed that the performance is nearly the same, which indicates the FD method can estimate the gradient with high accuracy under low dimensional states.

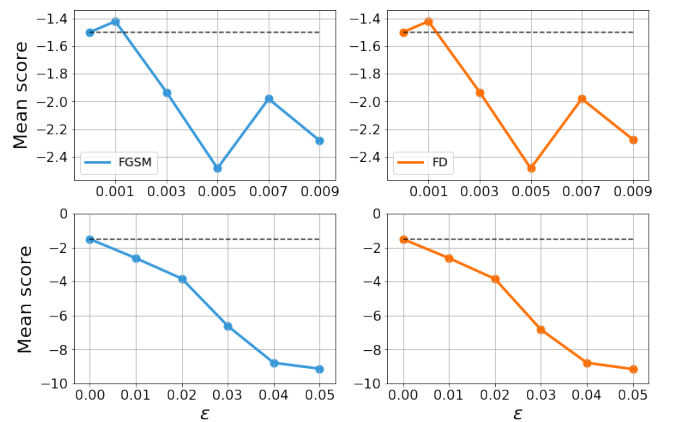

Fig. 8. Performance of the DQN under FGSM and FD method

Fig. 9 and Fig. 10 shows the performance of adversarial attacks exploiting the transferability property when the attacker cannot query the target agent freely. For the transfer across policies case, adversarial examples are generated by a DQN with different parameter weights. For the transfer across algorithms case, adversarial examples are generated by an IQN with different NN structure, hyperparameters and training process. Moreover, the IQN is a distributional RL algorithm so that the process of calculating the gradient $\nabla_{s_t} J(\theta, s_t, y)$ is also different from the standard DQN. It can be shown that although all the three methods under these two cases are less effective compared with the white box condition, they all outperform the random noises in most cases, which indicates the transferability property holds true in this EMS problem. Also, the results match our expectation: the more we know about the target agent, the better the adversarial examples can be generated by another agent. The adversarial examples generated by the IQN is less effective than a DQN, which is more similar to the target agent.

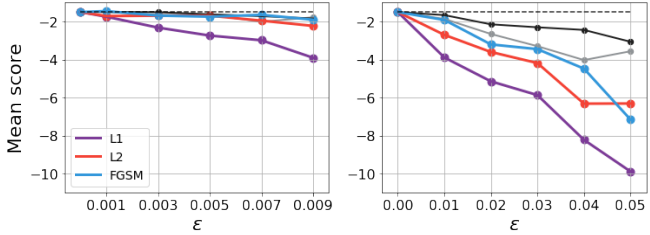

Fig. 9. Transfer across policies.

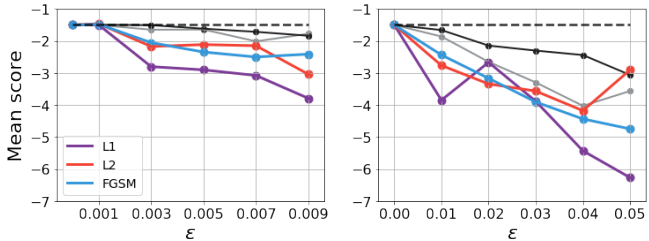

Fig. 10. Transfer across algorithms.

To provide some concrete examples of how the DQN agent behaves under adversarial attacks, the detailed trip information under attacks (Fig. 11 and 12) for the two previously shown trips in section II (Fig. 4 and 5) are shown. For the shorter trip, although no fuel is needed, 2.65 L fuel is consumed under the adversarial attacks. For the longer trip, some amount of fuel should be used at the last part of the trip to prevent the SOC from being lower than 10%. However, under the perturbed states, instead of increasing the value of $L_{set}$, the DQN agent chooses the opposite actions.

To further understand the impacts of adversarial examples on the DQN agent, we choose state $A$ and state $B$ from the two trips to show the comparison of action-values under normal states and perturbed states. In Fig. 13. it is clear that the agent will choose the action of decreasing the $L_{set}$ by 10 if the state information is correct, and the preference of actions are also easy to recognize according to the ranking of action-values. However, the differences between the action-values become unclear under the perturbed state. For state $B$ in Fig. 14, it can be shown that if the state information is transmitted correctly, the agent will consider that all actions will lead to bad results as all the action-values are very low, and the action of $+10$ is the best it can do. Nevertheless, when the state is perturbed, the magnitude of the action-values all change significantly and the ranking seems random.

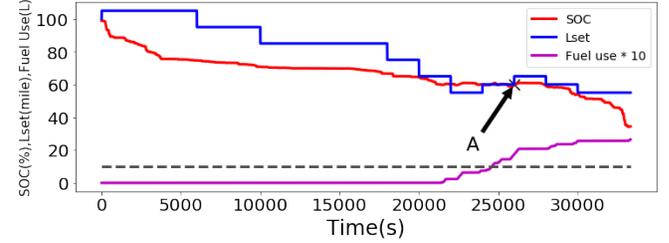

Fig. 11. Performance of DQN agent on the recorded trip shown in Fig. 4. under adversarial examples generated by $L_2$ method with $\varepsilon = 0.05$.

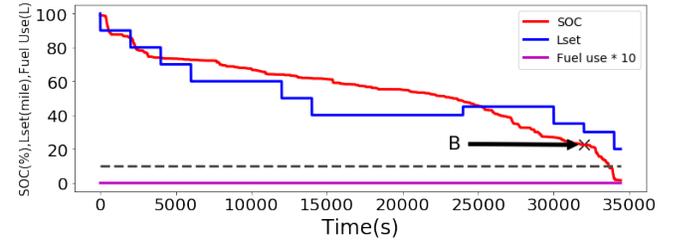

Fig. 12. Performance of DQN agent on the recorded trip shown in Fig. 5. under adversarial examples generated by FGSM method with $\varepsilon = 0.02$.

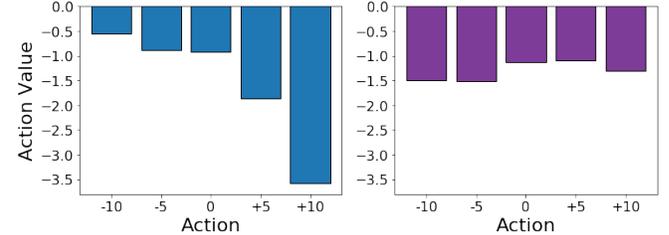

Fig. 13. Action-values comparison for state $A$. The left is for the original state and the right is for the perturbed state.

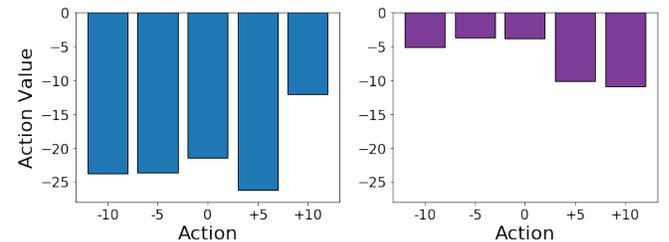

Fig. 14. Action-values comparison for state $B$. The left is for the original state and the right is for the perturbed state.

## V. DISCUSSION

It should be emphasized that the results reported from Fig.6 - Fig.10 are all mean scores over all the 52 trips. Consequently, all the used methods to generate adversarial examples are not guaranteed to degrade the performance of all the trips. Two directions to improve the adversarial attacks are briefly discussed below.

All the introduced and used methods in this work comes from the area of CV and ignores the fact that the inputs at each time step are correlated temporally in a RL task. Although these methods can degrade the performance of the target agent significantly, more efficient attacks can be achieved if the temporal correlation is utilized. To exploit this property, an adversarial agent can be trained to generate adversarial examples. This can be formulated as a new RL problem which considers the environment and the target agent combined as a new environment. The goal of the adversarial agent is to decrease the cumulated reward of the target agent in the original environment. By designing the reward function of the new problem, the norm and frequency of attacks can be controlled. It is expected that with a well-trained adversarial agent, the same level of performance drop can be reached with smaller perturbation and less frequent attacks. In [17] and [18], two heuristic and easy to implement methods to reduce the attack frequency by using the value function have been discussed.

Another downside of the used methods is no target action or target state is specified. The goal of the designed attacks is to change the action so that it is different from what the target agent intends to do. This can be partly addressed by more sophisticated and time-consuming methods like combining a generative model and a planning algorithm [18] and iterative least-likely class method [15].

There are at least two directions to build a more robust agent against possible adversarial attacks. First, detection methods can be utilized to monitor the inputs. Second, adversarial training can be applied [20][22].

## VI. Conclusion

In this work, the impacts of adversarial examples generated by several fast methods under white box and black box conditions on a DRL-based EMS are investigated. It is shown that adversarial examples can degrade the performance of the target agent significantly on average compared with the random noise. As DRL is more and more popular in the area of ITS, and robustness is one of the most important factors need to be considered when dealing with real-world problems, adversarial attacks should be considered carefully especially for safety-critical problems.

## Acknowledgment

The information, data, or work presented herein was funded in part by the Advanced Research Projects Agency-Energy (ARPA-E) U.S. Department of Energy, under Award Number DE-AR0000795. The views and opinions of authors expressed herein do not necessarily state or reflect those of the United States Government or any agency thereof.